\documentclass[runningheads,a4paper]{llncs}
\usepackage{amssymb}
\usepackage{graphicx}
\usepackage{amssymb}
\usepackage[utf8]{inputenc}
\usepackage{url}
\usepackage{float}
\usepackage{amsmath}
\usepackage{graphicx}
\usepackage{wrapfig}
\usepackage{booktabs}
\usepackage{multirow}
\usepackage{lipsum}

\makeatletter
\def\Hline{%
\noalign{\ifnum0=`}\fi\hrule \@height 1pt \futurelet
\reserved@a\@xhline}
\makeatother

\begin{document}

\title{Hibikino-Musashi@Home\\2022 Team Description Paper}

\author{
Tomoya Shiba
\and Tomohiro Ono
\and Shoshi Tokuno
\and Issei Uchino
\and Masaya Okamoto
\and Daiju Kanaoka
\and Kazutaka Takahashi
\and Kenta Tsukamoto
\and Yoshiaki Tsutsumi
\and Yugo Nakamura
\and Yukiya Fukuda
\and Yusuke Hoji
\and Hayato Amano
\and Yuma Kubota
\and Mayu Koresawa
\and Yoshifumi Sakai
\and Ryogo Takemoto
\and Katsunori Tamai
\and Kazuo Nakahara
\and Hiroyuki Hayashi
\and Satsuki Fujimatsu
\and Akinobu Mizutani
\and Yusuke Mizoguchi
\and Yuhei Yoshimitsu
\and Mayo Suzuka
\and Ikuya Matsumoto
\and Yuga Yano
\and Yuichiro Tanaka
\and Takashi Morie
\and Hakaru Tamukoh
}
\authorrunning{Tomoya Shiba et al.}
\institute{
Graduate school of life science and systems engineering,\\
Kyushu Institute of Technology,\\
2-4 Hibikino, Wakamatsu-ku, Kitakyushu 808-0196, Japan,\\
\email{hma@brain.kyutech.ac.jp} \\
\url{http://www.brain.kyutech.ac.jp/~hma/wordpress/}
}

\maketitle


\begin{abstract}
Our team, Hibikino-Musashi@Home (HMA), was founded in 2010.
It is based in Japan in the Kitakyushu Science and Research Park. Since 2010, we have annually participated in the RoboCup@Home Japan Open competition in the open platform league (OPL).
We participated as an open platform league team in the 2017 Nagoya RoboCup competition and as a domestic standard platform league (DSPL) team in the 2017 Nagoya, 2018 Montreal, 2019 Sydney, and 2021 Worldwide RoboCup competitions.
We also participated in the World Robot Challenge (WRC) 2018 in the service-robotics category of the partner-robot challenge (real space) and won first place.
Currently, we have 27 members from nine different laboratories within the Kyushu Institute of Technology and the university of Kitakyushu.
In this paper, we introduce the activities that have been performed by our team and the technologies that we use.
\end{abstract}


\section{Introduction}
Our team, Hibikino-Musashi@Home (HMA), was founded in 2010, and we have been competing annually in the RoboCup@Home Japan Open competition in the open platform league (OPL).
Our team is developing a home-service robot, and we intend to demonstrate our robot in this event in 2020 to present the outcome of our latest research.

In RoboCup 2017 Nagoya, we participated both in the OPL and the domestic standard platform league (DSPL) and in the RoboCup 2018 Montreal and RoboCup 2019 Sydney we participated in the DSPL.
Additionally, in the World Robot Challenge (WRC) 2018, we participated in the service-robotics category of the partner-robot challenge (real space).

In the RoboCup 2017, 2018, 2019, 2021 competitions and in the WRC 2018,  2020 we used a Toyota human support robot (HSR) \cite{toyota_hsr}. We were awarded the first prize at the WRC 2018, 2020, the second prize at the RoboCup 2021 and third prize at the RoboCup 2019.

In this paper, we describe the technologies used in our robot.
In particular, this paper outlines our object recognition system that uses deep learning \cite{hinton2006fast}, improves the speed of HSR, and has a brain-inspired artificial intelligence model, which was originally proposed by us and is installed in our HSR.


\section{System overview}
Figure \ref{fig:softOverview} presents an overview of our HSR system.
We have used an HSR since 2016.
In this section, we will introduce the specifications of our HSR.

\subsection{Hardware overview}
We participated in RoboCup 2018 Montreal and 2019 Sydney with this HSR.
The computational resources built into the HSR were inadequate to support our intelligent systems and were unable to extract the maximum performance from the system.
To overcome this limitation, using an Official Standard Laptop for DSPL that can fulfill the computational requirements of our intelligent systems has been permitted since RoboCup 2018 Montreal.
We use an MSI (Intel Core i9-11980HK CPU, 32GB RAM, and RTX-3080 GPU).
Consequently, the computer equipped inside the HSR can be used to run basic HSR software, such as sensor drivers, motion planning and, actuator drivers.
This has increased the operational stability of the HSR.
%

\subsection{Software overview}
\begin{figure}[bt]
\begin{center}
\includegraphics[scale=0.5]{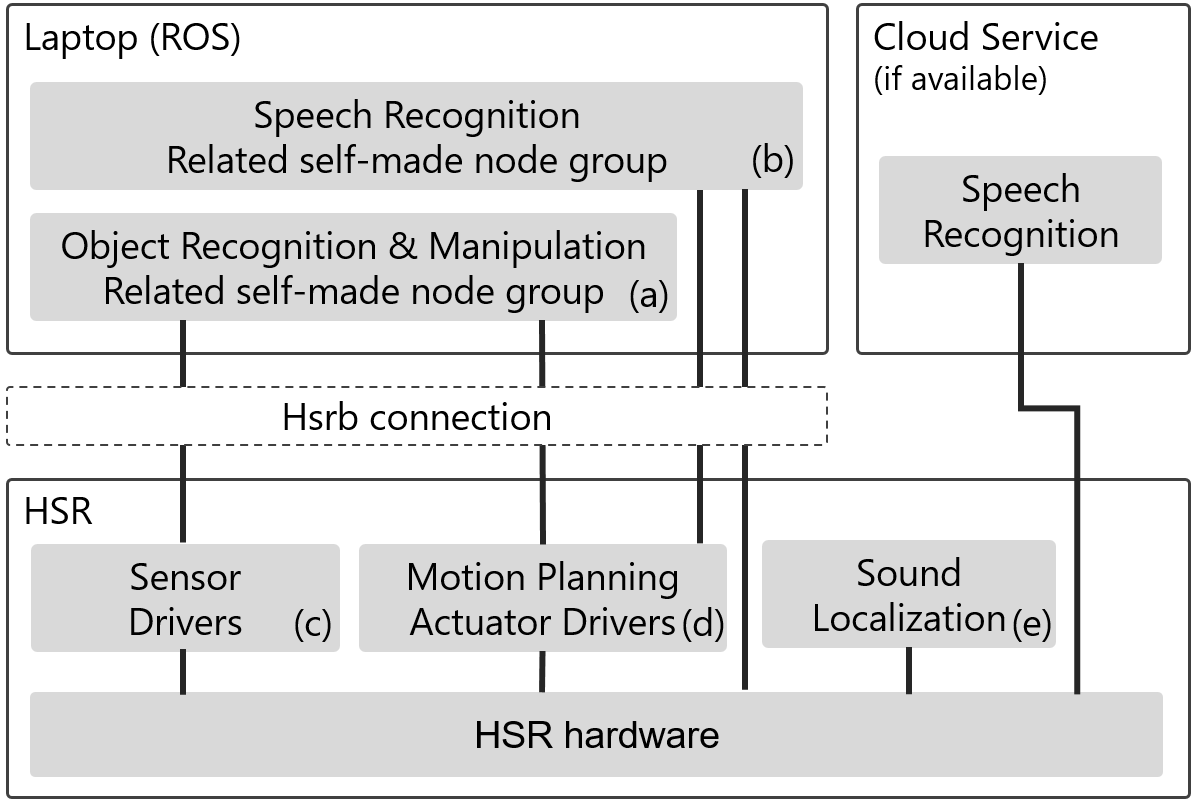}
\caption{Block diagram overview of our HSR system. [HSR, human-support robot; ROS, robot operating system]}
\label{fig:softOverview}
\end{center}
\end{figure}
In this section, we introduce the software installed in our HSR.
Figure \ref{fig:softOverview} shows the system installed in our HSR.
The system is based on the Robot Operating System \cite{ros}.
In our HSR system, laptop computer and a cloud service, if a network connection is available, are used for system processing.
The laptop is connected to a computer through an Hsrb interface.
The built-in computer specializes in low-layer systems, such as HSR sensor drivers, motion planning, and actuator drivers, as shown in Fig. \ref{fig:softOverview} (c) and (d).
Furthermore, the built-in computer has a sound localization system that use HARK \cite{hark}, as shown in Fig. \ref{fig:softOverview} (e).


\section{Object recognition}
In this section, we explain the object recognition system (shown in Fig. \ref{fig:softOverview} (a)), which is based on you look only once (YOLO) \cite{redmon2016you}.

To train YOLO, a complex annotation phase is required for annotating labels and bounding boxes of objects.
In the RoboCup@Home competition, predefined objects are typically announced during the setup days right before the start of the competition days.
Thus, we have limited time to train YOLO during the competition, and the annotation phase impedes the use of the trained YOLO during the competition days.

\begin{figure}[b]
\begin{center}
\includegraphics[scale=0.7]{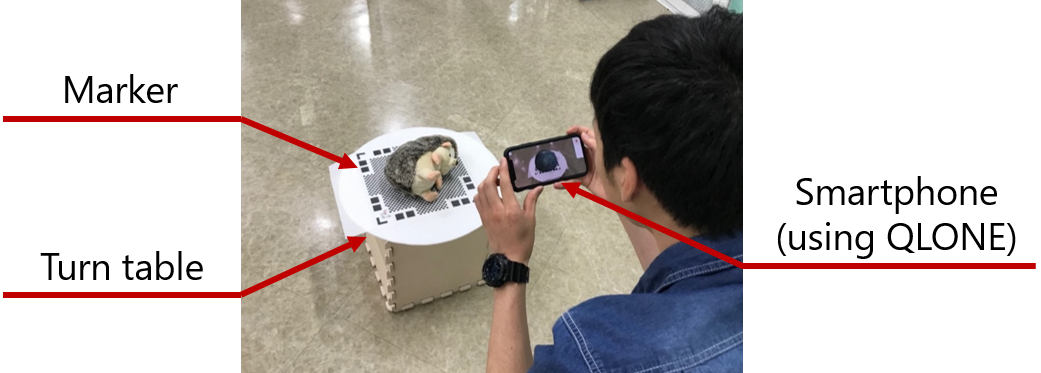}
\caption{Overview of proposed autonomous annotation system for YOLO.}
\label{fig:annotation1}
\end{center}
\end{figure}

We utilize an autonomous annotation system for YOLO using a three-dimensional (3D) scanner.
Figure \ref{fig:annotation1} shows an overview of the proposed system.
In this system, QLONE \cite{qlone}, a smartphone application capable of 3D scanning, is used.
QLONE makes it easy to create 3D models by placing objects on dedicated markers and shooting them.
We placed the marker and object on a turntable and created a 3D model.
In this method, the bottom surface of the object could not be shoot; thus, two 3D models can be created for each object by acquiring the flipped upside-down object.

\begin{figure}[tb]
\begin{center}
\includegraphics[scale=0.5]{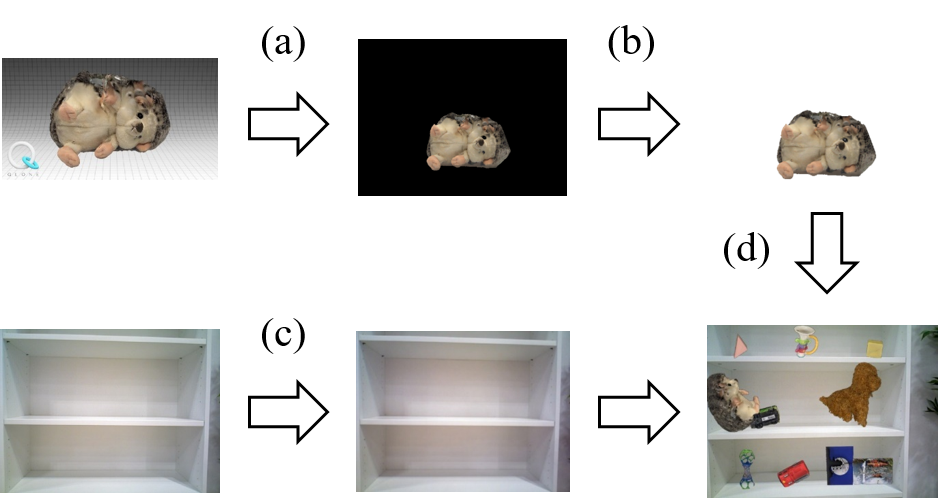}
\caption{Processing flow for generating training images for YOLO.}
\label{fig:annotation2}
\end{center}
\end{figure}

Figure \ref{fig:annotation2} shows the processing flow to generate training images for YOLO.

Multi-viewpoint images are automatically generated from the created two 3D models (Fig. \ref{fig:annotation2} (a)).
Then, we remove image backgrounds (Fig. \ref{fig:annotation2} (b)).

For backgrounds of the training images, we shoot background images, for example, a table, shelf, and other items.

To adapt to various lighting conditions, we apply the automatic color equation algorithm \cite{RIZZI20031663} to the background images (Fig. \ref{fig:annotation2} (c)).
To incorporate the object images into the background images, we define 20-25 object locations on the background images (the number of object locations depends on the background images).
Then, by placing the object images on the defined object locations autonomously, the training images for YOLO are generated (Fig. \ref{fig:annotation2} (d)).
If there are 15 class objects and 306 background images, 400,000 training images are generated.
Additionally, annotation data for the training images are generated autonomously because object labels and positions are known.

Image generation requires ~15 min (using six CPU cores in parallel), and training of YOLO requires approximately 6 h when using the GTX1080 GPU on a standard laptop.
Even though the generated training data are artificial, recognition of YOLO in actual environments works.
The accuracy when learning 10,000 epochs is 60.72\% in a mean average precision (mAP) evaluation.


\section{High-speed behavioral synthesis}
We are working to improve the speed of HSR from two viewpoints: behavioral synthesis and software processing speed.

Regarding behavioral synthesis, we reduce the wasted motion by combining and synthesizing several behaviors for the robot.
For instance, by moving each joint of the arm during navigation, the robot can move to the next action such as grasping without wasting any time as soon as the robot reaches an interim target location. 

Regarding the processing speed, we aim to operate all software at 30 Hz or higher.
To reduce the waiting time for software processing, which causes the robot to stop, the essential functions of the home service robot, such as object recognition and object grasping-point estimation, need to be executed in real time.
We optimized these functions for the Tidy Up Here task. 

\begin{figure}[bt]
  \begin{center}
    \includegraphics[scale=0.65]{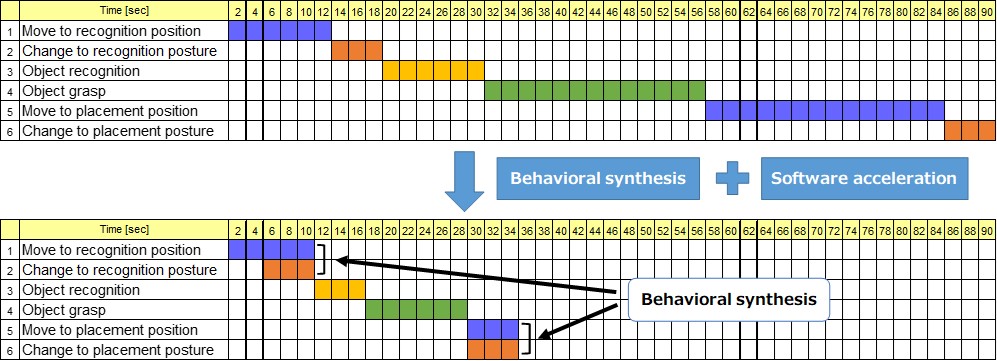}
    \caption{Speed comparison between a conventional system and the proposed high-speed system.}
    \label{fig:synthesis}
  \end{center}
\end{figure}

We used two optimized methods for that task in the WRC 2018 (Fig. \ref{fig:synthesis}). 
In the WRC 2018 results, for which we won first place, our achieved speedup was approximately 2.6 times the prior record. Our robot can tidy up within 34 s per object; thus, so we expect that it can tidy up approximately 20 objects in 15 min. 

\section{Unknown recognition}
In this section, we explain open set recognition for home service robots\cite{kanaoka2021visapp}.

Since the domestic environment in which a home service robot is supposed to work is subject to many changes, it is likely that the robot will always encounter unknown objects that it has not learned. In RoboCup, besides predefined objects are typically announced during the setup days right before the start of day, there are unknown objects that are not shown to the participants. It is very difficult to recognize unknown objects as unknown and to deal with them correctly, and only a few teams have tried to do so in past RoboCups.

\begin{figure}[tb]
\begin{center}
\includegraphics[scale=1.0]{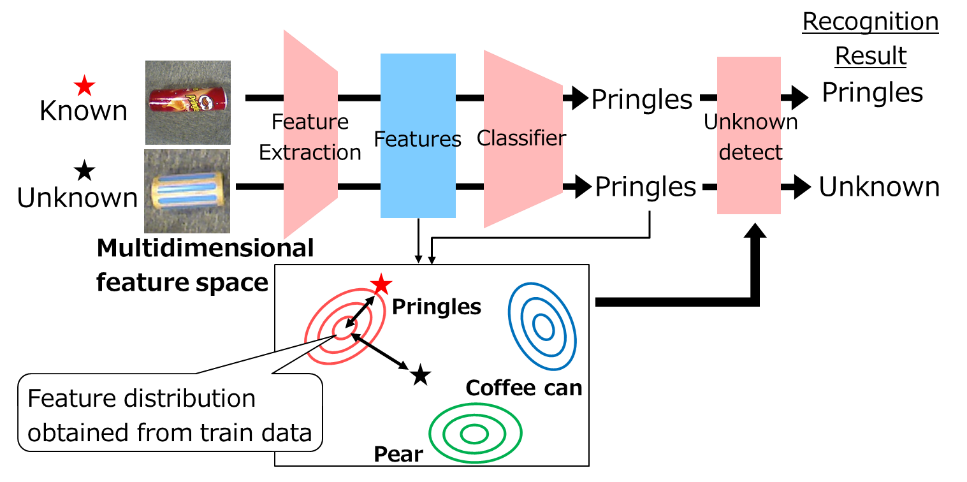}
\caption{Unknown recognition.}
\label{fig:unknownRecog}
\end{center}
\end{figure}

Open-set recognition is a recognition method that can detect unknown objects while performing general object recognition. We propose an open set recognition method that focuses on the multi-dimensional feature space formed in the middle layer of the neural network. Figure \ref{fig:unknownRecog} shows the proposed method. First, the model is trained using a back propagation. Next, training data is input to the trained neural network, and multivariate Gaussian fitting is performed for each class of clusters appearing in the multidimentional feature space. During inference, we first input a cropped object image and estimate the labels. Next, the Mahalanobis distance between the multivariate Gaussian distribution of the estimated labels and the input data is calculated. If the Mahalanobis distance exceeds the threshold value, the inference results are modified as unknown.

This system was implemented on the HSR at WRS2020. As a result, the system successfully recognized some of the unknown objects as unknown in the competition. Since WRS2020, we have also improved the input image preprocessor and the network to further improve the recognition performance.

\vspace{-0.3cm}
\section{Conclusions}
In this paper, we summarized the available information about our HSR, which we entered into RoboCup 2021 Worldwide.
The object recognition and improved speed of the HSR that we built into the robot were also described.
Currently, we are developing many different pieces of software for an HSR that will be entered into RoboCup 2022 Bangkok.

\vspace{-0.3cm}
\section*{Acknowledgment}
This work was supported by Ministry of Education, Culture, Sports, Science and Technology, Joint Graduate School Intelligent Car \& Robotics course (2012-2017),
Kitakyushu Foundation for the Advancement of Industry Science and Technology (2013-2015),
Kyushu Institute of Technology 100th anniversary commemoration project : student project (2015, 2018-2019) and YASKAWA electric corporation project (2016-2017),
JSPS KAKENHI grant number 17H01798 and 19J11524,
and the New Energy and Industrial Technology Development Organization (NEDO).

\newpage
\bibliography{ref}
\bibliographystyle{unsrt}

\newpage



\section*{Appendix 1: Competition results}

\begin{table}[t]
\begin{center}
\caption{Results of recent competitions. [DSPL, domestic standard-platform league; JSAI, Japanese Society for Artificial Intelligence; METI, Ministry of Economy, Trade and Industry (Japan); OPL, open-platform league; RSJ, Robotics Society of Japan]}
\label{tab:result}
\begin{tabular}{l|l|l} \hline
	\multicolumn{1}{c|}{Country} & \multicolumn{1}{c|}{Competition} & \multicolumn{1}{c}{Result} \\ \hline \hline

	Japan & RoboCup 2017 Nagoya & {\bf @Home DSPL 1st} \\
        				      && @Home OPL 5th \\ \hline

        Japan & RoboCup Japan Open 2018 Ogaki & @Home DSPL 2nd \\
        				                      && @Home OPL 1st \\
                                        	         && JSAI Award \\ \hline

        Canada & RoboCup 2018 Montreal & {\bf @Home DSPL 1st} \\
                                         && P\&G Dishwasher Challenge Award \\ \hline

        Japan & World Robot Challenge 2018 & {\bf Service Robotics Category} \\
                                              && {\bf Partner Robot Challenge Real Space 1st} \\
                                         	 && METI Minister's Award, RSJ Special Award \\ \hline

        Australia & RoboCup 2019 Sydney & @Home DSPL 3rd \\ \hline

        Japan & RoboCup Japan Open 2019 Nagaoka & @Home DSPL 1st \\
        				                          && @Home OPL 1st \\ \hline
        Japan & RoboCup Japan Open 2020 & @Home Simulation Technical Challenge 2nd \\
                                        && @Home DSPL 1st \\
                                        && @Home DSPL Technical Challenge 2nd \\
                                        && @Home OPL 1st \\
                                        && @Home OPL Technical Challenge 1st \\
                                        && @Home Simulation DSPL 2nd \\ \hline
        Worldwide & RoboCup Worldwide 2021 & @Home DSPL 2nd \\
                                        && \textbf{@Home Best Open Challenge Award 1st} \\
                                        && \textbf{@Home Best Test Performance: } \\
                                        && \textbf{    Go, Get It! 1st} \\
                                        && \textbf{@Home Best Go, Get It! 1st} \\ \hline
        Japan & World Robot Challenge 2020 & \bf{Service Robotics Category} \\
                                        && \bf{Partner Robot Challenge Real Space 1st} \\ \hline
        Japan & RoboCup Asia-Pacific 2021 Aichi Japan & \textbf{@Home DSPL 1st}\\
                                        && \textbf{@Home OPL 1st} \\ \hline
\end{tabular}
\end{center}
\end{table}

Table \ref{tab:result} shows the results achieved by our team in recent competitions.
We have participated in the RoboCup and World Robot Challenge for several years, and as a result, our team has won prizes and academic awards. \par
Notably, we participated in the RoboCup 2019 Sydney using the system described herein.
We were able to demonstrate the performance of HSR and our technologies.
Thanks to these results, we were awarded the third prize in that competition.

\section*{Appendix 2: Link to Team Video, Team Website}

\begin{itemize}
  \item Team Video \\ \url{https://www.youtube.com/watch?v=0loNuukvOec} \\
  \item Team Website \\ \url{http://www.brain.kyutech.ac.jp/~hma/} \\
  \item GitHub \\ \url{https://github.com/hibikino-musashi-athome} \\
  \item Facebook \\ \url{https://www.facebook.com/HibikinoMusashiAthome/} \\
  \item YouTube \\ \url{https://www.youtube.com/channel/UCJEeZZiDXijz6PidLiOtvwQ}
\end{itemize}

\section*{Appendix 3: Robot's Software Description}
For our robot we are using the following software:

\begin{itemize}
	\item OS: Ubuntu 18.04.
	\item Middleware: ROS Melodic.
	\item State management: SMACH (ROS).
	\item Speech recognition (English):
		\begin{itemize}
			\item rospeex \cite{rospeex}.
			\item Web Speech API.
			\item Kaldi.
		\end{itemize}
	\item Morphological Analysis Dependency Structure Analysis (English): SyntaxNet.
	\item Speech synthesis (English): Web Speech API.
	\item Speech recognition (Japanese): Julius.
	\item Morphological Analysis (Japanese): MeCab.
	\item Dependency structure analysis (Japanese): CaboCha.
	\item Speech synthesis (Japanese): Open JTalk.
	\item Sound location: HARK.
	\item Object detection: point cloud library (PCL) and you only look once (YOLO) \cite{redmon2016you}.
	\item Object recognition: YOLO.
	\item Human detection / tracking:
		\begin{itemize}
			\item Depth image + particle filter.
			\item OpenPose \cite{cao2017realtime}.
		\end{itemize}
	\item Face detection: Convolutional Neural Network.
	\item SLAM: rtabmap (ROS).
	\item Path planning: move\_base (ROS).
\end{itemize}

\end{document}